\def\year{2023}\relax
\pdfoutput=1
\documentclass[letterpaper]{article} 
\usepackage{aaai22}  
\usepackage{times}  
\usepackage{helvet}  
\usepackage{courier}  
\usepackage[hyphens]{url}  
\usepackage{graphicx} 
\usepackage{natbib}  
\usepackage{caption} 
\DeclareCaptionStyle{ruled}{labelfont=normalfont,labelsep=colon,strut=off} 
\usepackage{algorithm}
\usepackage{algorithmic}
\usepackage{multirow}
\usepackage{amssymb}
\usepackage{blindtext}
\usepackage{hyperref}
\usepackage{booktabs}
\usepackage{amsmath}
\usepackage{color}
\usepackage[final]{pdfpages}

\definecolor{Note_color}{rgb}{0.0, 0.0, 1.0}

\usepackage{newfloat}
\usepackage{listings}
\floatstyle{ruled}
\newfloat{listing}{tb}{lst}{}
\floatname{listing}{Listing}

\definecolor{sw_c}{rgb}{1.0, 1.0, 1.0}

\title{MIA-Former: Efficient and Robust Vision Transformers \\ via Multi-grained Input Adaptation}
\author{
    Zhongzhi Yu\textsuperscript{\rm 1},
    Yonggan Fu\textsuperscript{\rm 1},
    Sicheng Li\textsuperscript{\rm 2},
    Chaojian Li\textsuperscript{\rm 1},
    Yingyan Lin\textsuperscript{\rm 1}
}
\affiliations{
    \textsuperscript{\rm 1} Department of Electrical and Computer Engineering, Rice University
    \textsuperscript{\rm 2} Alibaba DAMO Academy

}

\usepackage{bibentry}

\begin{document}

\maketitle

\begin{abstract}

Vision transformers (ViTs) have recently demonstrated great success in various computer vision tasks, motivating a tremendously increased interest in their deployment into many real-world IoT applications. 
However, powerful ViTs are often too computationally expensive to be fitted onto real-world resource-constrained devices, due to (1) their quadratically increased complexity with the number of input tokens and (2) their overparameterized self-attention heads and model depth. In parallel, different images are of varied complexity and their different regions can contain various levels of visual information, e.g., a sky background is not as informative as a foreground object in object classification tasks, indicating that treating all regions/tokens equally in terms of model complexity is unnecessary while such opportunities for trimming down ViTs' complexity have not been fully explored. 
To this end, we propose a \textbf{M}ulti-grained \textbf{I}nput-\textbf{A}daptive Vision Trans\textbf{Former} framework dubbed \textbf{MIA-Former} that can input-adaptively adjust the structure of ViTs at three coarse-to-fine-grained granularities (i.e., model depth and the number of model heads/tokens).
In particular, our MIA-Former adopts a low-cost network trained with a hybrid supervised and reinforcement training method to skip unnecessary layers, heads, and tokens in an input adaptive manner, reducing the overall computational cost. Furthermore, an interesting side effect of our MIA-Former is that its resulting ViTs are naturally equipped with improved robustness against adversarial attacks over their static counterparts, because MIA-Former's multi-grained dynamic control improves the model diversity similar to the effect of ensemble and thus increases the 
difficulty of adversarial attacks against all its sub-models.
Extensive experiments and ablation studies validate that the proposed MIA-Former framework can (1) effectively allocate computation budgets adaptive to the difficulty of input images, achieving state-of-the-art (SOTA) accuracy-efficiency trade-offs, e.g., 20\% computation savings with the same or even a higher accuracy compared with SOTA dynamic transformer models, and (2) boost ViTs' robustness accuracy under various adversarial attacks over their vanilla counterparts by 2.4\% and 3.0\%, respectively. 
\end{abstract} 

\section{Introduction}
Vision transformers (ViTs) have been proven to be a powerful architecture on various computer vision tasks~\cite{dosovitskiy2020image, touvron2021training, chen2021pre, caron2021emerging, strudel2021segmenter}, especially when being scaled up with larger model sizes and more training data~\cite{dosovitskiy2020image, steiner2021train, ridnik2021imagenet}. 
However, powerful ViTs often come with prohibitive computational overhead.
Specifically, (1) the tokens consisting of merely background information and (2) redundant heads within the multi-head self-attention (MSA) module of ViTs which learn similar features can lead to unnecessary yet non-negligible inference costs.
Taking the widely used DeiT-Small model~\cite{touvron2021training} as an example, running inference on a single image with a resolution of $224\times224$ requires over 4.6 Giga floating point operations (GFLOPs), making it challenging to deploy ViTs onto many real-world resource-constrained devices for supporting intelligent internet of things (IoT) applications. Thus, there is an urgent need to reduce the computational cost of ViTs. 

On the other hand, in real-world applications, the complexity of images can vary significantly. As such, processing all the images with the same model complexity of ViTs could be overcooked. For example, for most of the time during video surveillance, the video may be just staring at an empty background, and the corresponding images can be processed with a naively simple model, saving a large portion of computational cost while still achieving satisfying accuracy. 
Thus, a straightforward solution for trimming down ViTs' complexity is to perform input-adaptive dynamic inference. 
Although dynamic inference has been extensively explored for convolutional neural networks (CNNs) through various dynamic dimensions (e.g., model depth, channel number, and model bit-width)~\cite{hu2020triple, wang2018skipnet, shen2020fractional, wang2020dual}, only a few pioneering works have considered this aspect for ViTs~\cite{rao2021dynamicvit, wang2021not}, which yet merely focus on reducing the computational budget by adaptively adjusting the number of input tokens.  
However, as suggested in~\cite{zhou2021deepvit}, the similarity between heads and feature maps can increase significantly in deeper ViT layers, implying that the token dimension is not the only source of redundancy, and the unexplored depth and head dimensions could lead to more efficient ViTs.

To this end, we target a multi-grained ViT framework that can fully explore the redundancy in ViTs and make the following contributions: 
\begin{itemize}
     \item We propose a \textbf{M}ulti-grained \textbf{I}nput-\textbf{A}daptive vision trans\textbf{Former} framework, dubbed MIA-Former, in order to trim down the redundancy of ViTs from multiple dimensions at three coarse-to-fine-grained granularities. 
    \item We propose a low-cost MIA-Controller to make input-adaptive decisions, which is jointly trained with the ViT models via a hybrid supervised and reinforcement learning (RL) scheme. 
    \item We empirically find that thanks to the proposed hybrid supervised and reinforcement training method, MIA-Former is equipped with improved robustness to various types of adversarial attacks, achieving a win-win in both robustness and efficiency. 
    \item Extensive experiments and ablation studies based on both DeiT-based~\cite{touvron2021training} and LeViT-based~\cite{graham2021levit} models show that the proposed MIA-Former can be used as a plug-in module on top of a wide range of ViTs to achieve better accuracy-efficiency trade-offs and boosted adversarial robustness, compared with state-of-the-art (SOTA) vanilla ViTs, input-adaptive ViTs, as well as CNNs. Specifically, MIA-Former achieves a 20.1\% FLOPs reduction and 2.4\% higher robustness accuracy under Projected Gradient Descent (PGD)~\cite{kurakin2016adversarial} attacks together with the same natural accuracy, compared with the original DeiT-Small model. 
\end{itemize}

\section{Related Works}

\textbf{Vision Transformers.}
Transformers are first introduced to natural language processing tasks in~\cite{vaswani2017attention} and achieve impressive performance in language modeling, machine translation, and other downstream tasks. It has been shown that the self-attention module in transformers can serve as an effective way to model the token-wise relationship of sentences. \cite{dosovitskiy2020image} then proposes ViTs, which is a pioneering work to extend transformer architectures to large scale compute vision tasks and matches SOTA CNNs' performance. Specifically,
ViTs first split an input image into a series of patches which are embedded into tokens before passing into the ViT blocks; 
Each ViT block consists of a stacked MSA and multi-layer perceptron (MLP) module to extract the global relationship among input tokens;
Multiple heads further enable ViT blocks to extract different features via different heads, improving the expressiveness of ViTs.
The success of ViTs on large scale image recognition tasks (e.g., ImageNet dataset~\cite{deng2009imagenet}) has inspired a series of following works to further exploit the expressive power of ViTs from different perspectives:~\cite{touvron2021training} performs an exhaustive search for the optimal training recipe for training ViTs; \cite{chen2021crossvit} exploits the impact of patch sizes on ViTs' achievable performance and proposes a parallel ViT architecture to simultaneously utilize information from both small and large patches to achieve better performance; and \cite{liu2021swin} proposes hierarchical structures for ViTs like ResNet~\cite{he2016deep}, and \cite{dong2021cswin} modifies the shape of patches from square to cross to balance the global and local attentions. 
Among them, LeViT~\cite{graham2021levit} achieves impressive performance by exploring the potential of applying convolutional layers before ViTs. Specifically, LeViT is the first ViT network that achieves a better accuracy-efficiency trade-off than EfficientNet~\cite{tan2019efficientnet} under certain FLOPs ranges.  


\textbf{Input-adaptive Inference.}
Adaptively activating different components of a deep neural network (DNN) in an input-dependent manner has been proved to be an effective way to reduce the inference cost, which has been widely explored for CNNs. Existing techniques in this regard can be roughly summarized into two categories: (1) making early predictions by introducing multiple side branch classifiers and dynamically exiting from one branch by analyzing the confidence of intermediate feature maps~\cite{kaya2019shallow, teerapittayanon2016branchynet} and (2) adaptively skipping certain components of the model, such as blocks, channels and even bit-width~\cite{wu2018blockdrop, wang2020dual, shen2020fractional, fu2020fractrain}. 
Nevertheless, the opportunities of input-adaptive inference have not yet been extensively explored in the scope of ViTs as existing works merely focus on dynamically adjust the input tokens, either by pruning out certain tokens~\cite{rao2021dynamicvit} or by adjusting the input patch size~\cite{wang2021not}. Other design dimensions in ViTs are still neglected, such as the number of attention heads and model depth, which play important roles in the overparameterization of ViTs.

\textbf{Adversarial Robustness and Model Efficiency.}
DNNs' robustness and efficiency are two critical features required in real-world applications. 
There are some pioneering works trying to simultaneously optimize both for CNNs. For example,
\cite{rakin2019robust, ye2019adversarial, sehwag2020hydra, fu2021drawing} combine pruning techniques with adversarial training methods. \cite{fu2021double} leverages the poor adversarial transferability between different precisions to win both robustness and efficiency, which can be accelerated by customized accelerators~\cite{fu20212} for further boosted efficiency. And \cite{rakin2018defend} uses dynamic quantization of activation functions to defend against adversarial examples; \cite{hu2020triple} introduces input-adaptive inference for simultaneously boosting robustness and efficiency. 
However, existing works have shown that CNNs and ViTs behave differently under adversarial attacks~\cite{shao2021adversarial,mao2021rethinking}, and how to win both adversarial robustness and model efficiency for ViTs is still an open question. 

\section{The Proposed Techniques}
In this section, we first present the MIA-Former framework which integrates a MIA-Controller module to dynamically control the activated subparts of ViTs in an input-dependent manner. 
After that, we introduce our proposed hybrid supervised and reinforcement training method that can effectively train  MIA-Former based ViTs to be equipped with both improved robustness and efficiency.

\begin{figure*}[t!]
    \centering
    \includegraphics[width=0.85\textwidth]{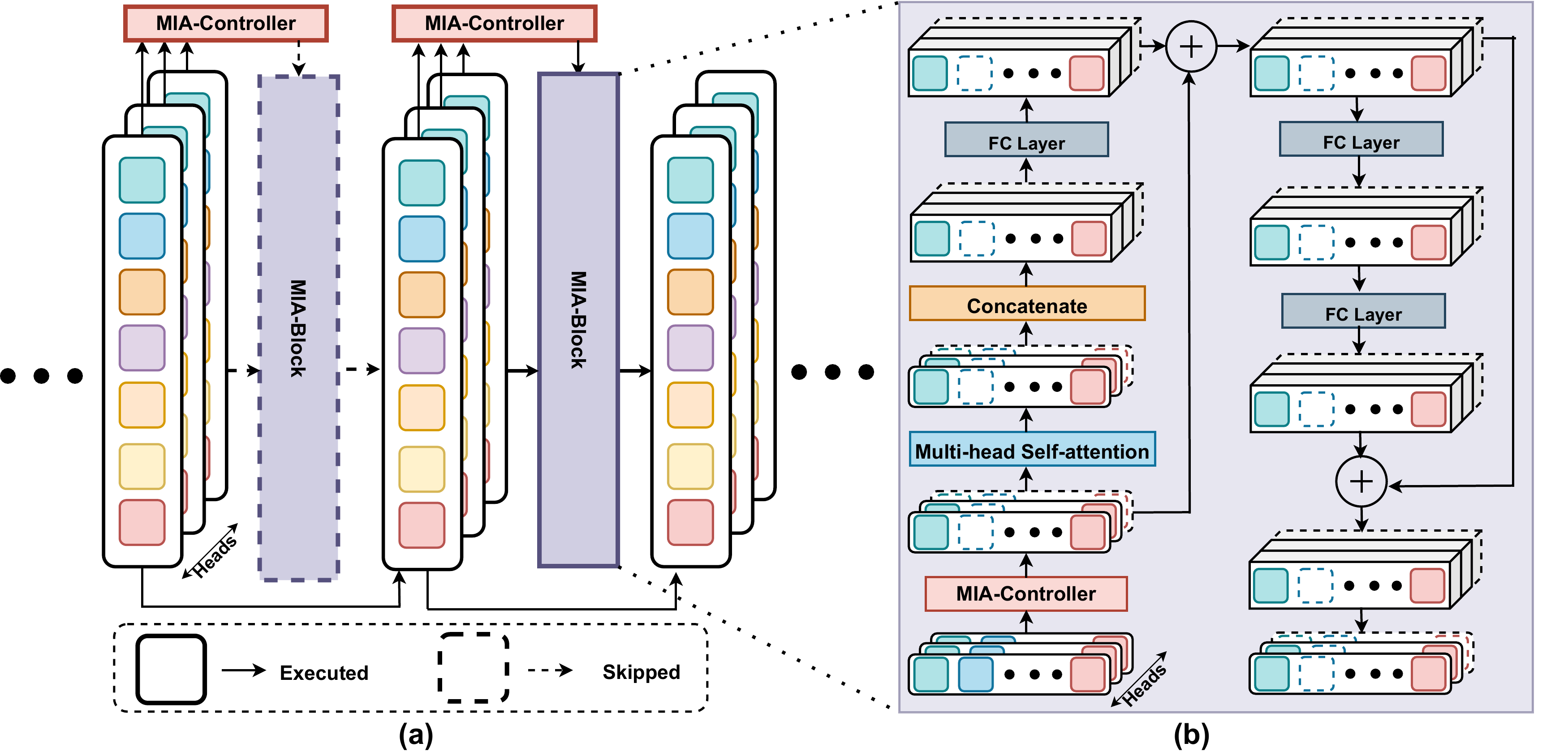}        
    \vspace{-0.5em}
    \caption{Overview of the proposed MIA-Former framework integrating a MIA-Controller and MIA-Block modules: (a) The MIA-Controller first decide whether to skip the whole upcoming MIA-Block, and (b) if the upcoming MIA-Block is not fully skipped, its MIA-Controller further dynamically masks out certain tokens and heads to deactivate the corresponding modules and thus reduce computational cost. }
    \label{fig:framework}
        \vspace{-1.2em}
\end{figure*}




\subsection{Overview}
Fig.~\ref{fig:framework} illustrates an overview of the proposed MIA-Former framework. 
On top of vanilla ViTs, MIA-Former adds a control module, dubbed MIA-Controller, to each backbone ViT block, MIA-Blocks. Specifically, MIA-Former integrates a MIA-Controller to first decide whether to mask out a given ViT block at a coarse granularity, as shown in Fig.~\ref{fig:framework}(a). In such cases, the outputs of a previous block skip the current block and are directly fed into the next MIA-Controller and MIA-Block. On the other hand, 
if the current MIA-Block is not skipped as a whole, MIA-Former further uses the corresponding MIA-Controller to dynamically mask out certain tokens and heads to deactivate the corresponding modules during inference in an input-dependent manner, as shown in Fig.~\ref{fig:framework}(b). 

In vanilla ViT blocks, reducing the number of heads can only lead to linear FLOPs reduction in MSA modules; moreover, the computational overhead in the upcoming MLP module cannot be reduced, which is often the computational bottleneck. Thus, we propose to further proportionally reduce both the input and output size of the upcoming fully connected (FC) layers in the MIA-Block module. In this way, MIA-Former can lead to linear FLOPs reduction in the MSA module and nearly quadratic FLOPs reduction in the following MLP module when skipping certain heads during inference, which can lead to large savings of FLOPs in MIA-Former. Note that a skip connection is also added to pass the information of masked heads and tokens to the outputs of the next block, avoiding the permanent loss of information.

\subsection{MIA-Controller}
The key characteristic of the proposed MIA-Former is to dynamically adjust its structure and thus complexity at three coarse-to-fine-grained granularities (i.e., model depth, number of heads, and number of tokens). To achieve optimal trade-offs between efficiency and accuracy, MIA-Former adopts a light-weight controller to adaptively generate masks for skipping the corresponding uninformative blocks/heads/tokens on top of its vanilla ViT backbone model. 
However, it is non-trivial to derive such finer-grained learnable masks due to the large skipping policy space. To tackle this problem, we propose a light-weight controller, i.e.,  MIA-Controller, to generate the masks and thus corresponding skipping policy. Specifically, MIA-Controller first examines whether a ViT block shall be completely skipped; if not, it will skip certain heads and tokens of this ViT block based on the input feature complexity. 
The merit of introducing the above block-wise skipping is that it eliminates the necessity of the additional efforts for computing head- and token-wise skipping policy. 

For each block $l$ in ViTs, we maintain three binary masks, $D^l_b\in \{0,1\}$, $D^l_h\in \{0,1\}^H$, and $D^l_n\in\{0,1\}^N$ for skipping blocks, heads, and tokens, respectively, where $H$ and $N$ denote the number of heads and tokens, respectively. In particular,
an MIA-Controller is inserted ahead of each block $l$ with the outputs of the previous MIA-Block $I^l\in\mathcal{R}^{N_h\times N_w\times(HE)}$ as the inputs, where $N=N_h N_w$ is the spatial dimension of the token array and $E$ is the hidden dimension of each head in the ViT. 
The MIA-Controller first passes $I^l$ through a two-layer CNN ($\text{CNN}_b$) with pooling to extract the features $F^l_b$ which are then passed through an FC layer ($\text{FC}_b$) with Gumbel softmax to generate $D^l_b$, deciding whether to skip the whole block. Such a skipping pipeline can be formulated as:  

\begin{equation}
\begin{aligned}
    F^l_b &= \text{CNN}_b(I^l) \in\mathcal{R}^{1\times 1\times HE'}, \\
    G^l_b &= \text{FC}_b(F^l_b) \in\mathcal{R},\\
    D^l_b &= \text{Round}(G^l_b)\in\{0,1\},
\end{aligned}
\end{equation}
where $E' < E$ is the hidden dimension of each head after the CNN with pooling in the MIA-Controller, and we set $E'=E/4$ in this work, $Round()$ is rounding operation that rounds the value to the nearest integer.

If a ViT block is not skipped, the MIA-Controller further generates a token mask $D^l_n$ and a head mask $D^l_h$ for determining the skipping policy for the corresponding block's heads and tokens, respectively. 
For $D^l_h$, an FC layer ($\text{FC}_{h1}$) accepts $F^l_b$ as its inputs to extract the head features $F^l_h$, and then another FC layer ($\text{FC}_{h2}$) with Gumbel softmax is adopted to generate the mask $D^l_h$:
\begin{equation}
    \begin{aligned}
        F^l_h &= \text{FC}_{h1}(F^l_b) \in\mathcal{R}^{H\times E''} \\
        G^l_h &= \text{FC}_{h2}(F^l_h) \in\mathcal{R}^{H} \\ 
        D^l_h &= \text{Round}(G^l_h)\in\{0,1\}^{H}.
    \end{aligned}
\end{equation}



For $D^l_n$, MIA-Former follows a similar procedure as \cite{rao2021dynamicvit}. 
Specifically, $I^l$ is reshaped to $I^{l'}\in\mathcal{R}^{N\times (HE)}$ and then applied to a two-layer MLP ($\text{MLP}_{n}$) to extract the features $F^l_n$. 
After that, an FC layer ($\text{FC}_{n}$) with Gumbel softmax is used to process $F^l_n$ and generate the masks $D^l_n$. 
\begin{equation}
    \begin{aligned}
        F^L_n &= \text{MLP}_n(I^{l'}) \in\mathcal{R}^{N\times HE'} \\
        G^l_n &= \text{FC}_n(F^l_n) \in\mathcal{R}^{N} \\
        D^l_n &= \text{Round}(G^l_n)\in\{0,1\}^{N}.
    \end{aligned}
\end{equation}

The featuremap $I^l_{F}$ after the MIA-Controller to the MIA-Block $l$ are computed as: 
\begin{equation}
    \begin{aligned}
    I^l_{F} = &I^l.\text{reshape}(N_h,N_w,H,E)\odot \\
    &D^l_b.\text{reshape}(1,1,1,1) \odot D^l_h.\text{reshape}(1,1,H,1)\odot \\
    &D^l_n.\text{reshape}(N_h, N_w, 1, 1),  
    \end{aligned}
\end{equation}
where $\odot$ represents the element-wise matrix multiplication (broadcast will be performed if the matrix shapes do not match).

\subsection{Hybrid Supervised and Reinforcement Training}
Given the complexity of MIA-Former, directly training all its components leads to unstable training and thus inferior performance. Thus, we propose a hybrid supervised and reinforcement training pipeline, which consists of three steps: (1) MIA-Controller pretraining, where we fix the parameters in the pretrained MIA-Block and pretrain the MIA-Controller until they keep activating all components of the MIA-Former without any skipping, 
(2) MIA-Former co-training, where we co-train the MIA-Block and the MIA-Controller with a hybrid loss function in a differentiable way, and (3) skipping policy finetuning with hybrid RL, where we finetune the MIA-Block and the MIA-Controller with a hybrid supervised and RL training method. We will illustrate the detail of the above stages in the remaining part of this section. 

\textbf{MIA-Controller Pretraining.}
As a randomly initialized MIA-Controller randomly skips different modules within the model, directly co-training the MIA-Block and MIA-Controller lead to inferior performance. 
We conjecture this is due to the fact that at the beginning of training, a large portion of the model is randomly skipped, leading to a significant deviation from the original learned distribution of pretrained ViTs. To tackle this, we propose to first pretrain the MIA-Controller with the MIA-Block weights fixed until it does not skip any components, and the whole MIA-Former should behave exactly the same as its backbone ViT at the end of this stage.
To achieve this, we train the MIA-Controller with the pretrain loss defined as: 
\begin{equation}
    \mathcal{L}_{pretrain} = \sum_{l=0}^{L}[(1-D^l_b)+(1-D^l_h)+(1-D^l_n)],
\end{equation}
where $L$ is the total number of blocks in a MIA-Former. 
When $\mathcal{L}_{pretrain}$ is minimized to zero, the MIA-Controller pretraining stage is finished. 

\textbf{MIA-Former Co-training.}
The learning objective of MIA-Former is to reduce the computational cost while preserving the model accuracy. 
To this end, we define the training loss as follow,
\begin{equation}
    \mathcal{L}_{diff} = \mathcal{L}_{task} + \alpha \mathcal{L}_{cost}, 
\end{equation}
where $L_{task}$ is the task loss, $L_{cost}$ is the computational cost loss, $\alpha$ is a weighted factor that trades off importance between accuracy and computational budget. In this work,
we define $L_{cost}$ as 
\begin{equation}
    L_{cost} = \frac{\text{FLOPs}_{exec}}{\text{FLOPs}_{total}},
\end{equation}
where $\text{FLOPs}_{exec}$ and $\text{FLOPs}_{total}$ are the total FLOPs of the executed parts in a MIA-Former and the total FLOPs when executing the whole MIA-Former model. 

To ensure that the MIA-Former can adaptively allocate computational budget among all the input samples, instead of setting $\alpha$ to a fixed value, we dynamically change the sign of $\alpha$ during training. Specifically, when $\text{FLOPs}_{exec}$ is larger than the given target FLOPs ($\text{FLOPs}_{target}$), we set $\alpha > 0$ to penalize MIA-Former for adapting to a smaller computation budget. On the other hand, when $\text{FLOPs}_{exec}$ is smaller than $\text{FLOPs}_{target}$, we set $\alpha < 0$ to encourage the model to execute more components in the MIA-Former. 

\begin{figure*}[]
    \centering
    \includegraphics[width=0.7\textwidth]{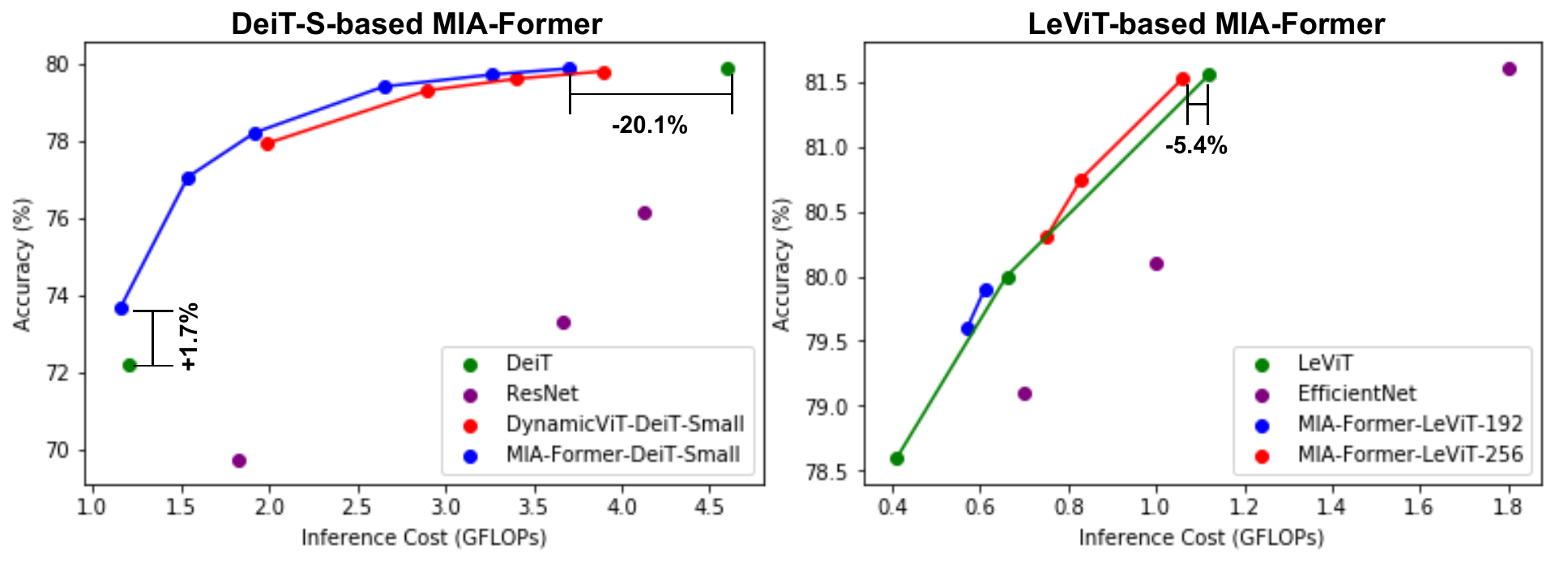}
            \vspace{-0.5em}
    \caption{Benchmarking model performance (top-1 accuracy) and inference efficiency (FLOPs) trade-off, where we compare our proposed MIA-Former with SOTA dynamic ViTs (DynamicViT) and other SOTA image classification ViTs/CNNs on the ImageNet dataset. }
    \label{fig:benchmark}
        \vspace{-1em}
\end{figure*}
\begin{figure*}[htb]
    \centering
    \includegraphics[width=0.7\textwidth]{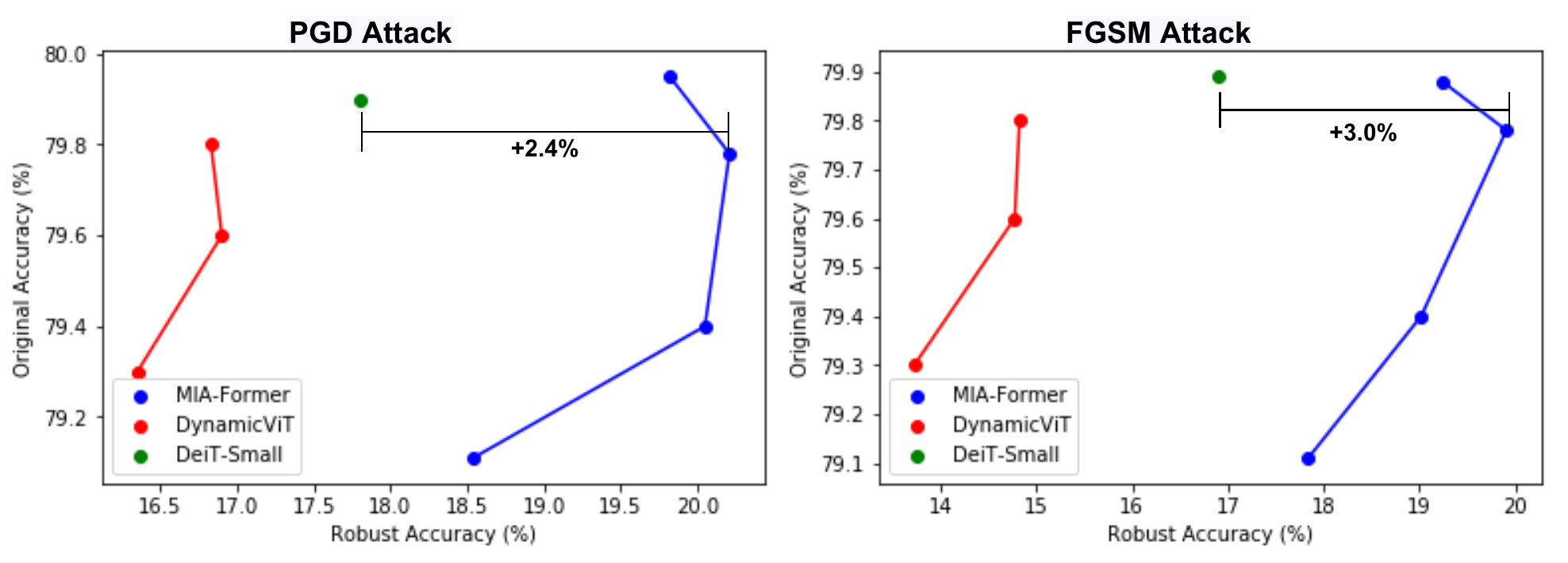}
        \vspace{-0.5em}
    \caption{Comparison of MIA-Former with DynamicViT and its backbone transformer models on the achieved ImageNet top-1 accuracy and robustness accuracy under both PGD and FGSM attacks. }
    \label{fig:robust-dynamic}
        \vspace{-1em}
\end{figure*}

\subsubsection{Skipping Policy Finetuning with Hybrid RL.}
We further finetune the MIA-Former with a hybrid RL method to achieve better performance. 
Specifically, we use an A2C-based~\cite{mnih2016asynchronous} RL learning method to train the MIA-Controller and MIA-Block modules in a differentiable manner. To make the MIA-Controller compatible with the above A2C algorithm, we replace the last FC layer in each of the skipping dimensions (i.e., depth, heads, and tokens) in an MIA-Controller with two parallel FC layers as the actor and critic network in the A2C algorithm, respectively, constructing a simple RL agent.  
To guarantee stable training, we further inherit all trained weights from the MIA-Block and 75\% of weights in the remaining parts of the MIA-Controller. In this way, the newly introduced RL agent can start from the learned features from the trained MIA-Controller, while the MIA-Block is also adapted to the dynamically skipping mechanism. 
The reward function of RL is defined as: 
\begin{equation}
    \mathcal{R} = Y + \beta(\text{FLOPs}_{target} - \text{FLOPs}_{exec})
\end{equation}
where $Y$ is a binary value indicating the task accuracy, and $\beta$ is a weighted factor that trades off ViTs' model accuracy and efficiency. 

To make the MIA-Block compatible with RL-based MIA-Controller, we also co-train the MIA-Block alone with MIA-Controller in this stage with the optimization objective defined as: 
\begin{equation}
 \min_{\omega} \mathcal{L}_{hybrid} = \mathcal{L}_{task} - \mathcal{R}, 
\end{equation}
where $\omega$ is the weights of the MIA-Block. 

\section{Experiment Results}

In this section, we first introduce the experiment setup, including models, datasets, training hyperparams, baselines, and evaluation metrics. Then, we validate the superiority of MIA-Former in boosting both efficiency and robustness. In particular, we final that (1) MIA-Former can boost the efficiency, i.e., achieving better accuracy vs. efficiency trade-off than both vanilla models and models with SOTA dynamic ViT techniques, and (2) MIA-Former can boost the robustness, i.e., achieving higher adversarial robustness while preserving the accuracy compared with both vanilla ViTs and dynamic ViT variants. Furthermore, we perform ablation study on (1) the comparison of the redundancy between different dimensions of MIA-Former based on the relative accuracy gap when running MIA-Former with different subsets of the input-adaptive granularities, and (2) which training strategy for learning the skipping policy can win better adversarial robustness in MIA-Former. Finally, we visualize the skipping policies on a subset of input samples as well as the skipping ratio distributions of different modules in the proposed MIA-Former.

\subsection{Settings}
\subsubsection{Models, Datasets, and Baselines.} We evaluate our proposed MIA-Former over \textbf{three ViT models} (i.e., DeiT-Small~\cite{touvron2021training}, LeViT-192 and LeViT-256~\cite{graham2021levit}) on ImageNet-1K dataset~\cite{deng2009imagenet}. We benchmark our method with vanilla ViTs, SOTA dynamic ViT method DynamicViT~\cite{rao2021dynamicvit} and other SOTA ViTs/CNNs designs. For adversarial robustness, we evaluate the proposed MIA-Former using Projected Gradient Descent (PGD)~\cite{madry2017towards} attack under $L_{Inf}$ constraint with a perturbation strength of 0.002 and Fast Signed Gradient Matching (FGSM) attack under $L_2$ constraint with a perturbation strength of 0.03~\cite{athalye2018obfuscated}, following the adversarial attack setting in~\cite{chen2020robust}.

\subsubsection{Training Recipe.} We adopt a three-stage training strategy: \underline{Stage 1: MIA-Controller pretraining}: we use an Adam~\cite{kingma2014adam} optimizer with a learning rate of 1e-4 to train the MIA-Controller with fixed MIA-Block until $\mathcal{L}_{pretrain}$ is decreased to 0. 
\underline{Stage 2: MIA-Former co-training}: we use an AdamW~\cite{loshchilov2017decoupled} optimizer with a batch size of 1024 and a learning rate of 1e-5/1e-3 to train the MIA-Block/MIA-Controller, respectively, for 200 epochs. We set the $\alpha$ to $0.1\times\frac{L_{cls}}{L_{cost}}$. 
\underline{Stage 3: Skipping policy finetuning with hybrid RL}: after inserting the RL agents, we first train the RL agent for 20 epochs with all other parameter fixed and then unfreeze other parameters and co-train the MIA-Former for a total of 50 epochs.  

\begin{table}[]
    \centering
    \caption{Comparing MIA-Former with SOTA ViTs/CNNs on ImageNet. We compare our MIA-Former with SOTA models under comparable FLOPs. We refer to MIA-Former with different backbones as MIA-Former-BACKBONE. We also include the results of vanilla models as references.}
    \resizebox{\linewidth}{!}{
    \begin{tabular}{ccc}
    \toprule[1pt]
        Model & GFLOPs & Accuracy (\%) \\
        \midrule
        MobileNet~\cite{howard2017mobilenets}            & 0.58 & 70.6 \\
        PVT-v2~\cite{wang2021pvtv2}                      & 0.61 & 76.9 \\  
        PiT-Ti~\cite{heo2021rethinking}               & 0.70 & 74.6 \\ 
        EfficientNet-B1~\cite{tan2019efficientnet}      & 0.70 & 79.1 \\ 
        LeViT-192~\cite{graham2021levit}            & 0.66 & 80.0 \\        IPE~\cite{chen2021exploring}                  & 0.88 & 78.6 \\ 
       \textbf{ MIA-Former-LeViT-192} & \textbf{0.61} &\textbf{ 79.9 }\\ 
        \midrule
        DeiT-Tiny~\cite{touvron2021training}         & 1.30 & 72.2 \\ 
        PVT-v2~\cite{wang2021pvtv2}
        Swin~\cite{liu2021swin}                 & 0.98 & 74.5 \\
        PiT-XS~\cite{heo2021rethinking}               & 1.40 & 79.1 \\ 
        LocalViT~\cite{li2021localvit}             & 1.05 & 74.9 \\
        CoaT~\cite{xu2021co}                 & 0.95 & 73.8 \\        EfficientNet-B2~\cite{tan2019efficientnet}      & 1.00 & 80.1 \\
        LeViT-256~\cite{graham2021levit}            & 1.12 & 81.5 \\
        \textbf{MIA-Former-LeViT-256} &\textbf{ 1.06} & \textbf{81.5} \\
        \bottomrule[1pt]
    \end{tabular}
    }
    \label{tab:main}
    \vspace{-1.1em}
\end{table}
\subsection{Benchmark with SOTA Designs}
\subsubsection{Enhanced accuracy-efficiency trade-off.}
We apply the MIA-Former framework on top of three ViT models, including DeiT-Small (one of the most widely used ViTs), LeViT-192, and LeViT-256, which are SOTA ViT models with the optimal accuracy-efficiency trade-off. As shown in Fig.~\ref{fig:benchmark}, we observe that the MIA-Former can achieve a 20.1\% reduction in FLOPs on top of vanilla DeiT-Small with comparable accuracy. 
We further benchmark the proposed MIA-Former with other SOTA CNNs and ViTs in Tab.~\ref{tab:main} and show that the proposed MIA-Former pushes forward the frontier of the achievable accuracy-efficiency trade-off with a 5.4\% reduction in FLOPs on top of LeViT-256 and a comparable accuracy.

\begin{figure}
    \centering
    \includegraphics[width=0.7\linewidth]{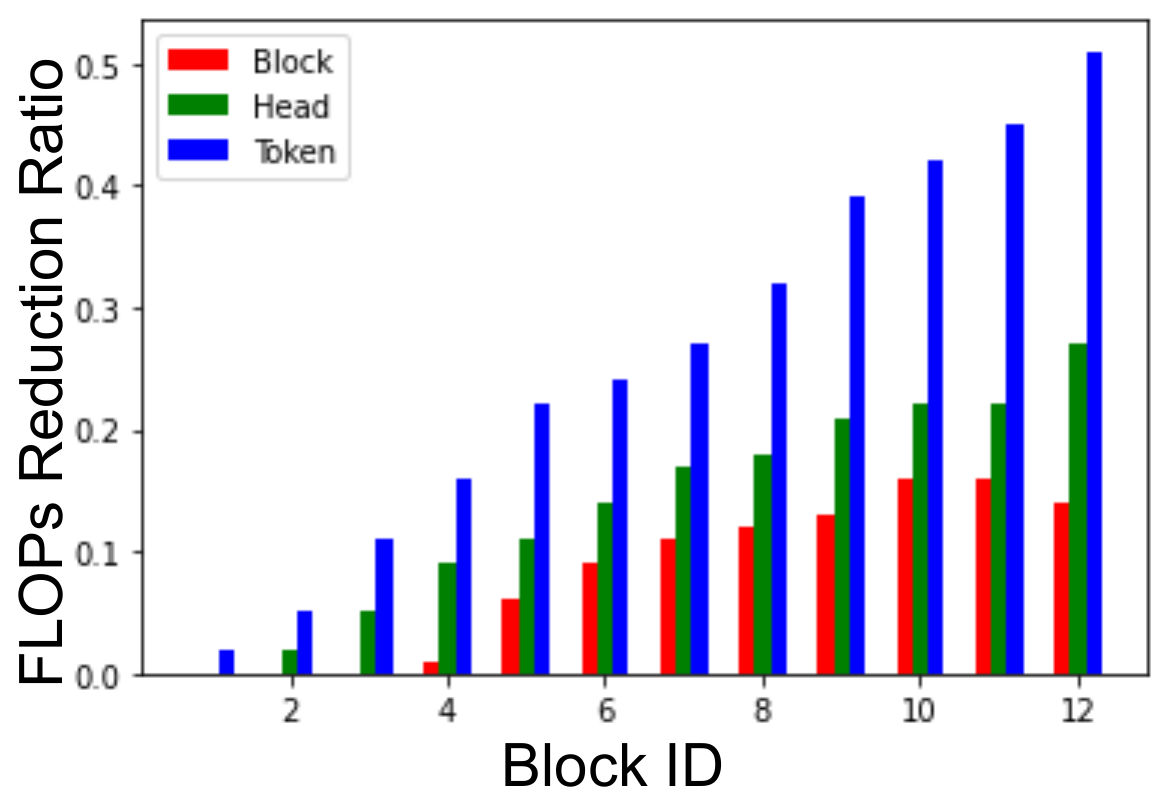}
    \vspace{-0.5em}
    \caption{Visualizing the blockwise skipping ratios along different dynamic dimensions on MIA-Former-DeiT-Small.}
    \vspace{-0.5em}
    \label{fig:skipping}
\end{figure}
\begin{figure*}[tb!]
    \centering
    \includegraphics[width=0.9\textwidth]{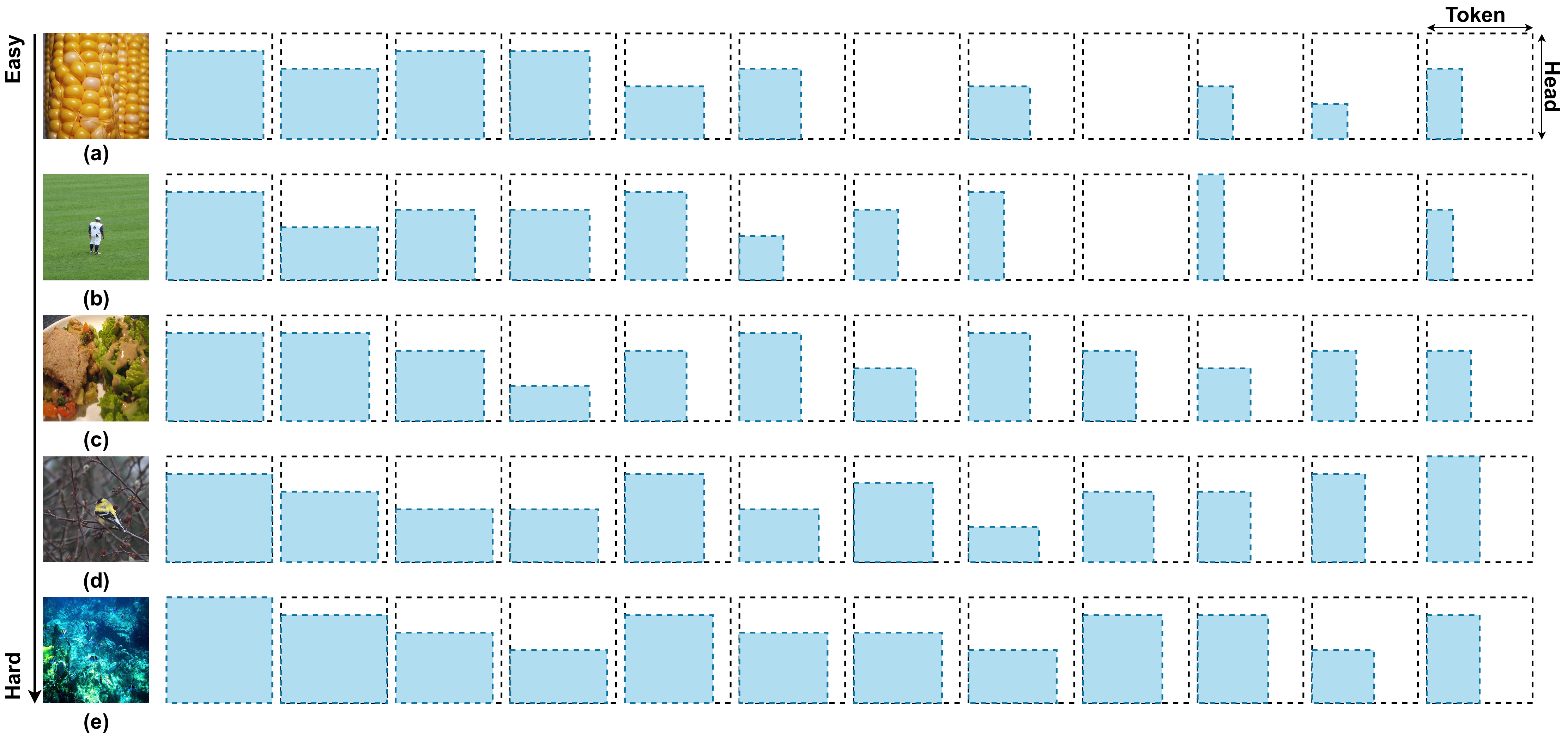}       
    \caption{Visualizing MIA-Former's input-adaptive skipping policy on different input samples. Each square in the figure represents a block in MIA-Former, the size of blue square indicates the executed part in the corresponding block, dashed square without any color represents fully skipped block. The height and width of the blue square indicate the executed number of heads and number of tokens, respectively.  }
    \label{fig:visualization}
    \vspace{-1.5em}
\end{figure*}

\subsubsection{Robustness improvement.}
We compare the adversarial robustness of  MIA-Former with vanilla DeiT-Small and DynamicViT~\cite{rao2021dynamicvit} models. As shown in Fig.~\ref{fig:robust-dynamic}, DynamicViT suffers from the reduction in robust accuracy in both cases of attacks, while our proposed method can achieve even higher model robustness compared with the original model, leading to a win-win in robustness and efficiency. Specifically, MIA-Former-DeiT-Small achieves up to a 2.4\%/3.0\% higher robust accuracy and a 26.1\% less FLOPs with comparable natural accuracy compared with the original DeiT-Small model under PGD/FGSM attacks, respectively. 



\begin{table}[tb]
    \centering
    \caption{Ablation study on different combination of dynamic dimensions.}
    \resizebox{0.8\linewidth}{!}{
    \begin{tabular}{ccc|cc}
    \toprule[1pt]
      \multicolumn{3}{c|}{Dynamic Dimension}  & \multirow{2}{*}{GFLOPs} & \multirow{2}{*}{Accuracy (\%)}\\
        Head & Depth & Token  \\ 
        \midrule
         &  &  & 4.6 & 79.9\\
        \checkmark &  &  & 3.9 & 78.7\\
         & \checkmark &  & 4.1 & 76.2\\
         &  & \checkmark & 3.8 & 79.7\\
        \checkmark & \checkmark &  & 4.0 & 78.6\\
        \checkmark &  & \checkmark & 3.7 & 79.8\\
         & \checkmark & \checkmark & 3.9 & 79.3\\
         \checkmark & \checkmark & \checkmark & \textbf{3.9} & \textbf{79.9} \\ 
         \bottomrule[1pt]
    \end{tabular}
    }
    \label{tab:dimension_redundancy}
\end{table}

\begin{table}[tb]
    \centering
            \vspace{-0.8em}
    \caption{Ablation study on the effect of hybrid training on model's robustness.}
    \resizebox{\linewidth}{!}{
   \begin{tabular}{c|c|cccc}
   \toprule[1pt]
    \multirow{2}{*}{} &\multirow{2}{*}{Target GFLOPs} &\multicolumn{4}{c}{Inherit Weight (\%)} \\
    \cmidrule{3-6}
    & &50 &75 &100 &Pretrain \\
    \midrule
    Clean Accuracy &\multirow{2}{*}{3.4} &79.2 &79.9 &79.9 &79.9 \\
    Robust Accuracy & &21.3 &19.8 &15.7 &14.2 \\
    \midrule
    Clean Accuracy &\multirow{2}{*}{3.9} &78.3 &79.6 &79.6 &79.5 \\
    Robust Accuracy & &22.8 &20.2 &17.5 &16 \\
    \bottomrule[1pt]
    \end{tabular}
    }
    \label{tab:RL ablation}
\end{table}

\subsection{Ablation Study}
\subsubsection{Analysis of the redundancy along each dimension.}
To better understand  the contribution of each dimension (i.e., head-wise, depth-wise, and token-wise) to the finally achieved accuracy-efficiency trade-off by MIA-Former, we conduct an ablation study by only enabling input-adaptive skipping at certain dimensions of MIA-Former on top of DeiT-Small. 
In Tab.~\ref{tab:dimension_redundancy}, the first row is the vanilla DeiT-Small model without any input-adaptive mechanism and the last row is the proposed MIA-Former with input-adaptive skipping enabled along all dimensions. 
As shown in Tab.~\ref{tab:dimension_redundancy}, by activating different dimensions, the performance of the MIA-Former varies significantly under similar inference FLOPs. 
Specifically, only activating dynamic depth suffers from $3.7\%$ lower accuracy with $0.2$ GFLOPs higher than the fully activated MIA-Former, indicating that the redundancy of the model in the depth-wise dimension is the lowest. 
On the other hand, even when only activating token-wise skipping, the accuracy only drops for $0.2\%$ with comparable FLOPs compared to MIA-Former, indicating the high redundancy in this dimension. 
This observation aligns with our intuition that depth-wise skipping is of the most coarse granularity while token-wise skipping has the finest granularity in ViTs.



\subsubsection{Effectiveness of finetuning the skipping policy with hybrid RL.}
How to inherit weights from the pretrained MIA-Controller when introducing the RL agent is critical to the finally achieved natural and robust accuracy. 
We study the impact of the portion of the inherited MIA-Controller weight at the beginning of the skipping policy finetuning stage on top of DeiT-Small under different FLOPs constraints.
As shown in Tab.~\ref{tab:RL ablation}, there exists a trade-off between the natural and robust accuracy when using different weight inheriting strategies. In particular,
inheriting a sufficient portion of pretrained MIA-Controller weights leads to significantly better natural accuracy. On the other hand, starting with a fully pretrained MIA-Controller without any reinitialization will degrade the robust accuracy. 
We suspect that this is because the inherited MIA-Controller makes the RL agent observe a similar feature as the differentiablly co-trained MIA-Controller in the previous training stage. The RL agent may try to make the decision in the same way as the differentiablly co-trained MIA-Controller. Thus, we pick a sweet spot from the trade-off and inherit 75\% weights throughout the paper.


\subsection{Skipping Policy Visualization}
We first summarize the statistical characteristic of the generated skipping policy on the validation set of ImageNet-1k~\cite{deng2009imagenet}. As in Fig.~\ref{fig:skipping}, we have the following observations: (1) deeper blocks have significantly higher redundancy than the shallower layers among all skipping dimensions in MIA-Former, and (2) the token-wise dimension has the highest skipping probability compared with the depth-wise and head-wise dimension, which is consistent with the ablation study on different combinations of skipping dimensions. 

To better understand the behavior of MIA-Former, we visualize the generated skipping policy of MIA-Former on different input samples. 
As shown in Fig.~\ref{fig:visualization}, MIA-Former generates different skipping policies based on different input samples. 
We can observe that MIA-Former can adaptively generate different policy based on the difficulty of input samples. For example, according to the comparison between Fig.~\ref{fig:visualization}(a) and (b), MIA-Former skips more from the token-wise dimension when processing (b) while skipping more from the head-wise dimension when processing (a) since (a) is full of corn in the image while (b) has a clean background with a small object in the front.


\section{Conclusion}
In this paper, we propose, develop, and validate MIA-Former, a multi-grained input-adaptive ViT framework, that is compatible with most of SOTA ViTs to achieve a higher accuracy-efficiency trade-off by dynamically skipping ViTs' blocks, heads, and tokens at coarse-to-fine granularities in an input-adaptive manner. 
The proposed hybrid supervised and reinforcement training method for effectively training the MIA-Former not only improves the achievable accuracy-efficiency trade-off of MIA-Former, but also boosts the robustness accuracy under different adversarial attacks, leading to a triple-win benefits. 
Extensive experiments and ablation studies show that the proposed MIA-Former achieves both efficiency and robustness improvement when being applied on top of various SOTA ViTs, pushing forward the frontier of ViTs' accuracy-efficiency trade-offs. To the best of our knowledge, this work is the first to simultaneously target multi-grained dynamic inference for ViTs.


\bibliography{main}

\end{document}